\documentclass[letterpaper, 10 pt, conference]{ieeeconf}  

\IEEEoverridecommandlockouts                              

\binoppenalty=\maxdimen
\relpenalty=\maxdimen

\usepackage{graphicx}
\usepackage{bm}
\usepackage{amsmath}
\usepackage{amssymb}
\usepackage{amsfonts}
\usepackage{mathtools} 
\usepackage{esvect}
\usepackage{textcomp}
\usepackage{booktabs}
\usepackage{tabularx}
\usepackage{float}
\usepackage{xcolor}
\usepackage[binary-units]{siunitx}
\usepackage{lipsum}
\usepackage{multirow}
\usepackage[algoruled, resetcount, linesnumbered ]{algorithm2e}

\SetAlgoSkip{SkipBeforeAndAfter}
\usepackage{algpseudocode}
\usepackage{dblfloatfix}

\usepackage{sidecap}

\usepackage{tikz}
\usetikzlibrary{fit,calc}

\usepackage{pgf} 
\usepackage{pgfplots}
\usepgfplotslibrary{external}
\usepackage{pgfplotstable}
\usepackage{pgfmath}
\usepackage{subcaption}

\graphicspath{{./figures/}}

\newcommand{\xx}{\bm{\mathrm{x}}} 
\newcommand{\ww}{\bm{\mathrm{u}}} 
\newcommand{\tx}{\bm{\mathtt{t}}^*}
\newcommand{\txx}{\mathtt{t_x}^*}
\newcommand{\txy}{\mathtt{t_y}^*}
\newcommand{\Rx}{\bm{\mathtt{R}}^*}
\newcommand{\Ax}{\bm{\mathtt{R}}}
\newcommand{\bx}{\bm{\mathtt{t}}}
\newcommand{\bxx}{\mathtt{t_x}}
\newcommand{\bxy}{\mathtt{t_y}}

\newcommand{\fvec}{\bm{\mathrm{f}}}
\newcommand{\gvec}{\bm{\mathrm{g}}}
\newcommand{\ff}{f}
\newcommand{\Ff}{\mathcal{F}}
\newcommand{\ft}{\hat{f}}
\newcommand{\gt}{\hat{g}}

\newcommand{\mbympp}{\texttt{f-mbym}}

\newcommand{\mbymppm}{\texttt{f-mbym@255}}
\newcommand{\mbympph}{\texttt{f-mbym@511}}
\newcommand{\mbym}{\texttt{mbym}}
\newcommand{\mbyml}{\texttt{mbym@127}}

\newcommand{\mbymh}{\texttt{mbym@511}}

\newcommand{\rawpp}{\texttt{f-raw}}

\newcommand{\raw}{\texttt{raw}}

\newcommand{\noentry}{-}
\newcommand{\maxpool}{\mathtt{MP}}
\newcommand{\feature}{\bm{\mathrm{h}}}
\newcommand{\bn}{\mathtt{BN}}
\newcommand{\relu}{\mathtt{Relu}}
\newcommand{\sigmoid}{\mathtt{Sigmoid}}
\newcommand{\upsample}{\mathtt{BL}}

\usepackage[acronym]{glossaries}
\newacronym{fmcw}{FMCW}{Frequency-Modulated Continuous-Wave}
\newacronym{ft}{FT}{Fourier Transform}
\newacronym{fft}{FFT}{Fast Fourier Transform}
\newacronym{fmt}{FMT}{Fourier-Mellin Transform}
\newacronym{vo}{VO}{Visual Odometry}
\newacronym{ekf}{EKF}{Extended Kalman Filter}
\newacronym{slam}{SLAM}{Simultaneous Localisation and Mapping}
\newacronym{ro}{RO}{Radar Odometry}
\newacronym{mbym}{MByM}{Masking by Moving}
\newacronym{fmbym}{f-MByM}{Fast Masking by Moving}
\newacronym{dnn}{DNN}{Deep Neural Network}
\newacronym{cnn}{CNN}{Convolutional Neural Network}
\newacronym{cpu}{CPU}{Central Processing Unit}
\newacronym{gpu}{GPU}{Graphical Processing Unit}
\newacronym{ndt}{NDT}{Normal Distributions Transform}
\newacronym{ram}{RAM}{Random Access Memory}
\newacronym{cuda}{CUDA}{Compute Unified Device Architecture}

\usepackage{cleveref}
\crefname{table}{Tab.}{Tabs.}
\crefname{figure}{Fig.}{Figs.}
\crefname{section}{Sec.}{Secs.}
\crefname{equation}{Eq.}{Eqs.}
\crefname{algorithm}{Alg.}{Algs.}

\usepackage{cuted}

\usepackage{scalefnt}

\renewcommand{\baselinestretch}{0.95}

\title{\large \bf
Fast-MbyM: Leveraging Translational Invariance\\of the Fourier Transform for Efficient and Accurate Radar Odometry
}

\author{Rob Weston$^{*}$, Matthew Gadd$^{\dagger}$, Daniele De Martini$^{\dagger}$, Paul Newman$^{\dagger}$, and Ingmar Posner$^{*}$ 
\\
$^{*}$Applied Artificial Intelligence Lab (A2I), $^{\dagger}$Mobile Robotics Group (MRG), University of Oxford
\\
{\tt\small \{robw, mattgadd, daniele, pnewman, ingmar\}@robots.ox.ac.uk}%
\\
{\tt \small https://github.com/applied-ai-lab/f-mbym}%
\vspace{-1em}
}

\begin{document}

\maketitle
\thispagestyle{empty}
\pagestyle{empty}

\begin{abstract}
\gls{mbym}, provides robust and accurate radar odometry measurements through an exhaustive correlative search across discretised pose candidates. However, this dense search creates a significant computational bottleneck which hinders real-time performance when high-end GPUs are not available. Utilising the translational invariance of the Fourier Transform, in our approach, \gls{fmbym}, we decouple the search for angle and translation. By maintaining end-to-end differentiability a neural network is used to mask scans and trained by supervising pose prediction directly. Training faster and with less memory, utilising a decoupled search allows f-MbyM to achieve significant run-time performance improvements on a CPU (\SI{168}{\percent)} and to run in real-time on embedded devices, in stark contrast to MbyM. Throughout, our approach remains accurate and competitive with the best radar odometry variants available in the literature -- achieving an end-point drift of \SI{2.01}{\percent} in translation and \SI{6.3}{\deg\per\kilo\metre} on the \textit{Oxford Radar RobotCar Dataset}.

\end{abstract}
\section{Introduction}
\label{sec:Introduction}
In recent years, \gls{ro} has emerged as a valuable alternative to lidar and vision based approaches due to radar's robustness to adverse conditions and long sensing horizon. However, noise artefacts inherent in the sensor imaging process make this task challenging.
The work of Cen and Newman \cite{cen2018precise} first demonstrated the potential of radar as an alternative to lidar and vision for this task and since then has sparked significant interest in \gls{ro}.

Whilst sparse point-based \gls{ro} methods such as \cite{cen2018precise,cen2019radar,aldera2019fast,burnett2021we,burnett2021radar,barnes2020under,adolfsson2021cfear} have shown significant promise, Barnes \textit{et al.}~\cite{barnes2019masking} recently established the benefits that a dense approach brings to this problem setting. By masking radar observations using a \acrshort{dnn} before adopting a traditional brute-force scan matching procedure, \mbym~\emph{learns} a feature embedding explicitly optimised for \gls{ro}.
As robust and interpretable as a traditional scan matching procedure, \mbym~was able to significantly outperform the previous state of the art~\cite{cen2018precise}.

However, as our experiments demonstrate, \mbym~in its original incarnation is unable to run in \emph{real-time} on a laptop at all but the smallest resolutions and not at all on an embedded device. The requirement for a high-end \acrshort{gpu} for real-time performance represents a significant hindrance for deployment scenarios where the cost or power requirements of such hardware is prohibitive.

In this work
we propose a number of modifications to the original \mbym~approach which result in significantly faster run-time performance, enabling real-time performance at higher resolutions on both \acrshortpl{cpu} and embedded devices. In particular, instead of performing a brute-force search over all possible combinations of translation and angle, we exploit properties of the \acrlong{ft} to search for the angle between the two scans \emph{independent} of translation. By adopting this decoupled approach, we significantly reduce computation. Our approach, \mbympp, retains end-to-end differentiability and thus the use of a \acrshort{cnn} to mask radar scans, learning a radar scan representation explicitly optimised for \gls{ro}. \mbympp, is shown in \cref{fig:teaser}. Like \mbym~our model is trained end-to-end in a supervised fashion. However, our modifications allow \mbympp~to be trained much more rapidly and with much less memory.


\begin{strip}
\centering
\includegraphics[width=0.95\textwidth]{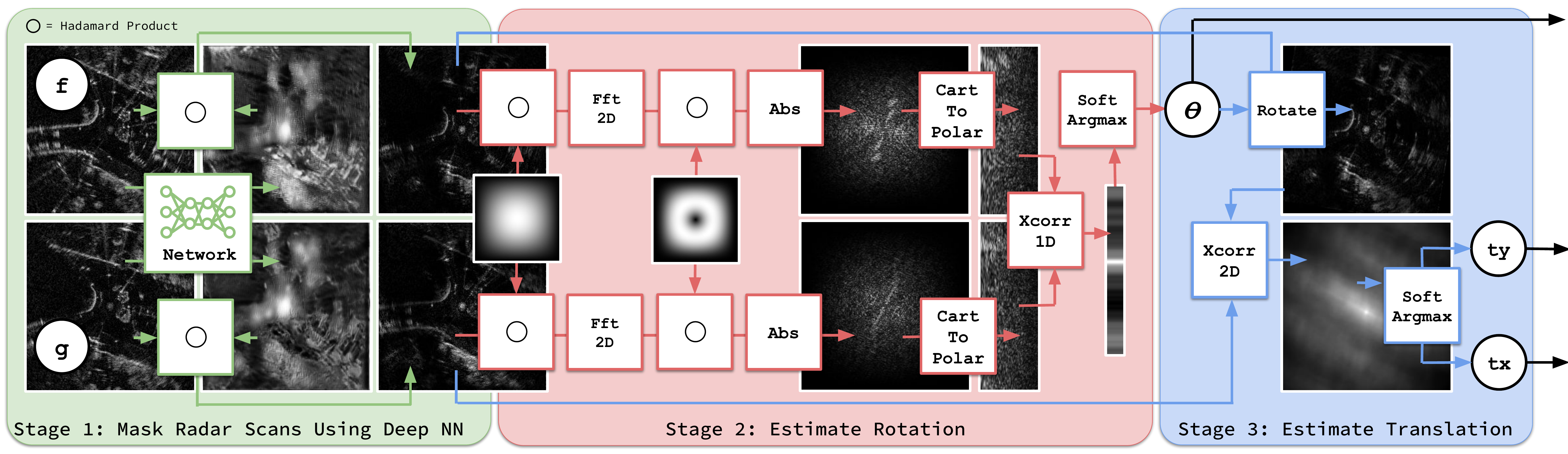}
\captionof{figure}{
Given radar scans $\bm{\mathrm{f}}$ and $\bm{\mathrm{g}}$ our system outputs the relative pose $[\theta, \bxx, 
\bxy]$ between them in three phases: (1) each radar scan is masked using a deep neural network; (2) the rotation $\theta$ is determined by maximising the correlation between the magnitude of their fourier transforms in polar co-ordinates; (3) the translation $[\bxx, \bxy]$ is determined by maximising the correlation between $\bm{\mathrm{f}}$ and $\bm{\mathrm{g}}$ rotated by the now known angle $\theta$. Using our approach we are able to determine $\theta$ independently of $[\bxx, \bxy]$ allowing us to achieve real-time performance on a CPU and embedded devices. Crucially, this entire procedure is end-to-end differentiable allowing us to explicitly optimise our network for radar pose estimation.
}
\label{fig:teaser}
\end{strip}

By providing a greater run-time efficiency at higher resolutions our best performing real-time model achieves an end-point error of \SI{2.01}{\percent} in translation and \SI{6.3}{\deg\per\kilo\metre} in rotation on the \textit{Oxford Radar RobotCar Dataset} \cite{barnes2020radarrobotcar}, outperforming the best real-time \mbym~model in accuracy whilst running $168\%$ faster on a \acrshort{cpu} and in real-time (at \SI{6}{\hertz}) on a \textit{Jetson} GPU. Our approach remains competitive with the current state-of the-art, point-based methods.
\section{Related Work}
\label{sec:RelatedWorks}
In recent years the work of Cen \textit{et al.} \cite{cen2018precise,cen2019radar} has demonstrated the potential of \gls{ro} as an alternative to vision and lidar, sparking a significant resurgence in interest in \gls{ro}. Cen and Newman \cite{cen2018precise} propose a global shape similarity metric to match features between scans whilst in their subsequent work \cite{cen2019radar} a gradient-based feature detector and a new graph matching strategy are shown to improve performance.

Since then several methods have been proposed \cite{barnes2019masking,barnes2020under,burnett2021radar,aldera2019fast,aldera2019could}, outperforming \cite{cen2018precise,cen2019radar} with gains attained through a combination of motion compensation \cite{burnett2021we,burnett2021radar,adolfsson2021cfear}, fault diagnosis and filtering~\cite{aldera2019could}, as well as new learnt \cite{barnes2020under,burnett2021radar,aldera2019fast} and rule-based \cite{adolfsson2021cfear} feature representations. In \cite{aldera2019fast}, as an alternative to hand-crafted feature extraction proposed in \cite{cen2018precise}, Aldera \textit{et al.} propose to extract temporally consistent radar feature points using a \acrshort{dnn}. In this approach labels for stable points are generated by accumulating a histogram of points across time and over wide baselines. Instead, Barnes and Posner \cite{barnes2020under} extract and learn radar feature representations by supervising pose prediction directly. This results in a significant reduction in end-point error when compared to \cite{cen2019radar}. In \cite{burnett2021we} Burnett \textit{et al.} find that motion compensating scans yields significant boosts in \gls{ro} performance (when compared to \cite{cen2018precise}). Combining this with an unsupervised adaptation of \cite{barnes2020under}, in \cite{burnett2021radar}, Burnett \textit{et al.} are able to slightly outperform \cite{barnes2020under} without requiring ground truth odometry measurements to train their system. In an alternative and recently proposed approach \cite{adolfsson2021cfear} a robust point-to-line metric is used, in combination with motion compensation and estimation over a sliding window of past observations.

In contrast to the sparse methods mentioned above, \acrlong{mbym} \cite{barnes2019masking} adopts a dense approach; using a correlative scan matching procedure in combination with a learnt feature space supervised for pose prediction, the optimal pose is searched for across a dense grid of candidates. Through this approach \mbym~is able to outperform sparse variants \cite{cen2018precise,cen2019radar,burnett2021radar,burnett2021we}. However, while a dense search results in excellent performance it comes with a significant computational cost. This cost may be offset when high-end modern graphical processing hardware is available as demonstrated by the timing results shown in \cite{barnes2019masking}, but means that \mbym~struggles to run online when the cost or power requirements of such hardware is prohibitive. The learnt element of \mbym~can also lead to geographical overfitting, where the model performs better in the areas it has been trained. In this work our aim is to tackle the former of these problems, noting that as larger scale and more varied radar odometry datasets become available models should become less prone to overfitting. Nonetheless, further investigation into combating geographical overfitting in the low data regime, remains an interesting area for future research.

Building upon \cite{barnes2019masking} we also adopt a dense scan matching procedure with a learnt feature representation supervised directly for pose estimation. However, we propose to overcome the computational burden of the dense search by decoupling the search for angle and translation between scans, exploiting the translational invariance of the Fourier Transform~\cite{lim1990two}. 
This property alongside the scale invariance property of the Mellin Transform (MT) are combined to form the \gls{fmt}~\cite{lim1990two}. The \gls{fmt} has been widely exploited for image registration~\cite{casasent1976position,fmlimreg} as well as for visual odometry~\cite{kazik2011visual,ho2008optical}.

In the radar domain, Checchin and G{\'e}rossier~\cite{checchin2010radar} proposed to use the FMT for \gls{ro} over a decade ago. More recently \cite{park2020pharao} proposes to use a similar approach in their \gls{ro} system. In contrast, as the scale between scans is known, in our own work we rely on \emph{only} the Fourier translation property. In contrast to \cite{checchin2010radar,park2020pharao} we propose to mask radar observations using a \acrshort{cnn}. Using a differentiable implementation of the decoupled scan matching procedure allows us to learn a radar feature representation supervised for pose prediction without resorting to hand-crafted filtering or feature extraction and results in superior performance.
\section{Approach}
\label{sec:Approach}
We begin by formulating the problem (\cref{sec:approach-formulation}) and discuss the limitations of a na\"ive correlative scan matching procedure (\cref{sec:approach-mbym}). Next we show how by using properties of the Fourier Transform we are able to more efficiently search for the optimum pose by decoupling the search for rotation and translation (\cref{sec:approach-ours-theory}). In \cref{sec:approach-ours-practice} we propose a discrete and differentiable implementation. Finally, to improve performance the radar scans are filtered using a \acrlong{dnn} (\cref{sec:approach-ours-learnt-embedding}) which is trained explicitly for pose prediction (leveraging the differentiability of our scan matching implementation).

\subsection{Problem Formulation}
\label{sec:approach-formulation}
Let the signals $\ff(\xx) \in \mathbb{R}$ and $g(\xx') \in \mathbb{R}$ denote radar power measurements in two coordinate systems $\xx, \xx' \in \mathbb{R}^2$ related by a rigid-body transformation $[\Rx | \tx] \in \mathbb{SE}(2)$
\begin{equation}
\label{eqn:xd}
    \xx' = \Rx \xx + \tx
\end{equation}
where $\Rx = \Rx(\theta^*) \in \mathbb{SO}(2)$ is a 2D rotation matrix parameterised by yaw $\theta^* \in [0, 2\pi]$ and $\tx = [\txx, \txy]^\top \in \mathbb{R}^2$ is a translational offset. In this case the radar power measurements $f(\xx)$ and $g(\xx')$ are related as $f(\xx) \approx g(\Rx \xx + \tx)$, where this relationship is only approximate due to appearance change between the two frames.
The aim of our approach is to estimate the pose $[\Rx |  \tx] \in \mathbb{SE}(2)$ between the two coordinate systems, given access to $f(\xx)$ and $g(\xx')$.

\subsection{Correlative Scan Matching}
\label{sec:approach-mbym}
In a \emph{correlative scan matching} approach (such as in \cite{barnes2019masking}) the optimum pose $[\Rx| \tx]$ is found by maximising the correlation between the two scans
\begin{equation}
\label{eqn:ccormax}
\Rx, \txx, \txy = \text{argmax}_{\Ax, \bxx, \bxy} (f \star g)(\Ax, \bxx, \bxy)
\end{equation}
where $(f \star g)(\Ax, \bxx, \bxy)$ is the \emph{cross-correlation} operation:
\begin{equation}
\label{eq:corr}
(f \star g)(\Ax, \bxx, \bxy) = \int_{\mathbb{R}^2}f(\xx) g(\Ax \xx + [\bxx, \bxy]^\top) d \xx \text{ .}
\end{equation}
The optimum pose is found through a brute force approach partitioning the space $\Ax, \bx \in \mathbb{SO}(2) \times \mathbb{R}^2$ into discrete and evenly spaced pose candidates $\Ax, \bxx, \bxy \in \{\Ax_i \}_{i=1}^{n_\theta} \times \{ \mathtt{x}_i\}_{i=1}^{n_x} \times \{ \mathtt{y}_i\}_{i=1}^{n_y}$ and choosing the pose that maximises correlation between $f$ and $g$.
However, searching over every possible combination of $\Rx$, $\txx, \txy$ creates a significant computational bottleneck hindering real-time performance when high-end compute is not available.

\subsection{Exploiting Translational Invariance of the \acrlong{ft} for Efficient Pose Estimation}
\label{sec:approach-ours-theory}
We therefore utilise properties of the \acrlong{ft} to search for $\Ax \in \{\Ax_i \}_{i=1}^{n_\theta}$ \emph{independently} of $\bx \in \{ \mathtt{x}_i\}_{i=1}^{n_x} \times \{ \mathtt{y}_i\}_{i=1}^{n_y}$.
This is the key to the efficiency of our approach. With the radar signals related as $f(\xx) = g(\Rx \xx + \tx)$ their \acrlongpl{ft} are related as 
\begin{align}
    \ft (\ww) &:= \Ff[ \ff(\xx) ] := a_f(\ww) e^{j\phi_f(\ww)} \\
    \hat{g}(\ww') &:=  \Ff[ g(\xx') ] := a_g(\ww') e^{j \phi_g (\ww')} \\
    \hat{f}(\ww) &= \gt(\Rx \ww)e^{2 \pi j {\tx}^\top \Rx \ww} \label{eqn:affineft}
\end{align}
(see proof in~\cref{app:appendix})
where $\Ff : \mathbb{R} \rightarrow \mathbb{C}$ denotes the one-sided 2D \acrlong{ft} and $\ww=[u_1, u_2]^\top \in\mathbb{R}^2$ is the spatial frequency.
Here, their magnitudes $a_f(\ww) := |\hat{f}(\ww)|$, $a_g(\ww') :=| \hat{g}(\ww')|$ differ \emph{only} by a rotation, $a_f(\ww)= a_g(\Rx \ww)$ and are \emph{independent} of $\tx$\footnote{This can intuitively be understood by noting that translating the original 2D signal does not change the overall frequency content, merely shifts it to a new location (resulting in a phase shift between the two signals)}. Exploiting this result, an efficient algorithm for determining the optimum pose $[\Rx | \tx]$ emerges:
\subsubsection{Determine $\Rx$}
Considering $a_f(\ww)$ and $a_g(\ww')$ in polar coordinates $\tilde{a}_f(\bm{\omega})$ and $\tilde{a}_g(\bm{\omega}')$, where $\bm{\omega}(u_1, u_2) = \left[\tan^{-1}(\frac{u_2}{  u_1}), \sqrt{u_1^2 + u_2^2} \right]$ is the polar representation of the 2D spatial frequency plane, the rotation between 
$\ww$ and $\ww'$ will manifest as a translation between $\bm{\omega}$ and $\bm{\omega}'$:
the angle $\theta$ between the two signals can therefore be recovered as,
\begin{equation}
\label{eq:angle_search}
\theta^* = \text{argmax}_{\theta} (\tilde{a}_f \star \tilde{a}_g)(\bm{\mathtt{I}}, \theta, 0)
\end{equation}
where $\bm{\mathtt{I}} = \text{diag}([1, 1])$ and $\text{argmax}_{\theta} (\tilde{a}_f \star \tilde{a}_g)$ is the correlation as per \cref{eq:corr} between the magnitudes of the two signals after mapping to polar coordinates.

\subsubsection{Determine $\tx$}
Once $\Rx=\Rx(\theta^*)$ is known we are able to recover $\tx = [\txx, \txy]^\top$ as,
\begin{equation}
\label{eq:trans_search}
\txx, \txy = \text{argmax}_{\bxx, \bxy} (f \star g)(\Rx, \bxx, \bxy) 
\end{equation}
where $g$ is rotated by the rotation solved for in the previous step.
Compared to the na\"ive approach, where this last step must be performed for every yaw candidate $\Ax \in \{ \Ax_i \}_{i=1}^{n_\theta}$, this reduces computation by a factor of $n_\theta$.

\subsection{Implementation}
\label{sec:approach-ours-practice}

\begin{algorithm}[h]
\footnotesize
\caption{Fourier Scan Matching Procedure} \label{alg:scan_match}
\SetKwProg{Fn}{function}{:}{}
\Fn{ScanMatch($\fvec$, $\gvec$, $n_\theta=733$, $\delta_\theta=\pi / 733$, $T_\theta=2$, $n_{xy}=255$, $\delta_{xy}=0.4$, $T_{xy}=1$)}{
\tcc{Determine the pose $[\Rx, \tx]$ between the two scans $\fvec, \gvec \in \mathbb{R}^{n_{xy} \times n_{xy}}$}
$\{\theta_k \} = Linspace(-\frac{1}{2}\delta_\theta n_\theta, \frac{1}{2}\delta_\theta n_\theta, n_\theta)$ \\
$ \{\mathtt{x}_i \} = Linspace(-\frac{1}{2}\delta_{xy} n_{xy}, \frac{1}{2}\delta_{xy} n_{xy}, n_{xy})$ \\
$ \{\mathtt{x}_j\} = Linspace(-\frac{1}{2}\delta_{xy} n_{xy}, \frac{1}{2}\delta_{xy} n_{xy}, n_{xy})$ \\
\tcc{Stage 1: Determine $\theta^*$}
$\bm{\mathrm{h}}_{Hann} = HanningFilter(Shape(\fvec))$ \\
$\fvec, \gvec = \bm{\mathrm{h}}_{Hann} \circ \fvec, \bm{\mathrm{h}}_{Hann} \circ \gvec$ \\
$\hat{\fvec}, \hat{\gvec} = FFT2d(\fvec), FFT2d(\gvec)$ \\
$\bm{\mathrm{h}}_{Band} = BandPassFilter(Shape(\fvec))$ \\
$\hat{\fvec}, \hat{\gvec} =\bm{\mathrm{h}}_{Band} \circ \hat{\fvec}, \bm{\mathrm{h}}_{Band} \circ \hat{\gvec}$ \\
$\bm{\mathrm{a}}_f, \bm{\mathrm{a}}_g = Abs(\hat{\fvec}), Abs(\hat{\gvec})$  \\
$\tilde{\bm{\mathrm{a}}}_f, \tilde{\bm{\mathrm{a}}}_g = Cart2Pol(\bm{\mathrm{a}}_f), Cart2Pol(\bm{\mathrm{a}}_g)$ \\
$\tilde{\bm{\mathrm{a}}}_f, \tilde{\bm{\mathrm{a}}}_g = WrapPad(\bm{\mathrm{a}}_f), WrapPad(\bm{\mathrm{a}}_g)$ \\
$\bm{c}_{\theta r} = iFFT2d(FFT2d(\tilde{\bm{\mathrm{a}}}_f) \circ FFT2d(\tilde{\bm{\mathrm{a}}}_g))$ \label{2dcorr:1} \\
$\bm{\mathrm{c}}_\theta = Mean(\bm{\mathrm{c}}_{\theta r}, \text{dim}=\text{'r'})$ \\
$\theta = SoftArgMax(\bm{\mathrm{c}}_\theta, T_\theta, \{\theta_k\} \})$ \label{softamax:1} \\
\tcc{Determine $[\txx, \txy]$}
$\gvec' = Rotate(\gvec, \theta^*)$ \\
$\fvec, \gvec' = ZeroPad(\fvec), ZeroPad(\gvec')$ \\
$\bm{\mathrm{c}}_{xy} = iFFT2d(FFT2d(\fvec) \circ FFT2d(\gvec'))$ \label{2dcorr:2} \\
$\bxx', \bxy' = SoftArgMax(\bm{\mathrm{c}}_{xy}, T_{xy}, \{\mathtt{x}_i \} \times \{\mathtt{y}_j \} )$ \label{softamax:2} \\
$\Ax' = BuildSO2(-\theta)$ \\
$\bxx, \bxy = MatMul(\Ax', [\bxx', \bxy'])$ \\
\Return{$\theta, \bxx, \bxy$}}
\end{algorithm}

Whilst the approach so far was developed for continuous signals $f(\xx)$ and $g(\xx')$ in reality we only have access to discrete sets of power measurements $\fvec \in \mathbb{R}^{n_x \times n_y}$ and $\gvec\in \mathbb{R}^{n_x \times n_y}$ measured at locations $\xx, \xx' \in \{ \mathtt{x}_i\}_{i=1}^{n_x} \times \{ \mathtt{y}_i\}_{i=1}^{n_y}$ (assumed to fall over an evenly spaced grid).
\cref{alg:scan_match} therefore gives a discrete approximation to the approach developed up to this point. The function $ScanMatch$ takes as input $\fvec$ and $\gvec$ and returns the estimated pose $[\Ax, \bxx, \bxy]$. A diagram of our approach is found in~\cref{fig:teaser}.

The 2D correlation operator defined in~\cref{eq:corr} is approximated in~\cref{alg:scan_match} by its discrete counterpart and is implemented as a multiplication in the Fourier domain using the highly efficient $FFT2d$ and inverse $iFFT2d$ (lines~\ref{2dcorr:1} and \ref{2dcorr:2}).
The $\text{argmax}$ operation in \cref{eq:angle_search,eq:trans_search} is replaced with a soft approximation $SoftArgMax$ in lines \ref{softamax:1} and \ref{softamax:2} to ensure that the scan matching procedure maintains end-to-end differentiability. Here, a temperature controlled softmax is applied to the 2D correlation scores before a weighted sum is performed over its coordinates. This property will be exploited in~\cref{sec:approach-ours-learnt-embedding} to \emph{learn} a radar embedding optimised for pose prediction.
It was found that applying specific filtering and padding strategies was important to ensure correct operation. A Hanning filter~\cite{oppenheim1999discrete} is applied before performing the 2D FFT of $\fvec$ and $\gvec$ to reduce boundary artefacts and a band-pass filter was applied thereafter to reduce the impact of uninformative low and high frequencies.
As the angular dimension in polar-coordinates is periodic, applying circular padding to the power spectra along the angular dimension ($WrapPad$ in \cref{alg:scan_match}) significantly reduces boundary artefacts; on the translational directions, instead, we padded the spectra with zeros ($ZeroPad$ in \cref{alg:scan_match}). 
The functions $Rotate$ and $Cart2Pol$ are implemented using bi-linear interpolation in a similar approach to \cite{jaderberg2015spatial}. The number of range readings is set to $n_{xy}$.

\subsection{Learnt Radar Embeddings For Improved Odometry}
\label{sec:approach-ours-learnt-embedding}
Central to the success of our approach was an assumption that $f(\xx) \approx g(\Rx \xx + \tx)$. Of course there are several reasons why this condition might not hold in practice: dynamic objects, motion blur, occlusion, and noise all result in a power field that fluctuates from one time-step to the next. 
To counteract this, in a similar approach to \cite{barnes2019masking}, we propose to mask the radar power returns using a neural network $h_\alpha$ to filter the radar scans before scan matching:
\begin{gather}
 [\bm{\mathrm{m}}_f,\bm{\mathrm{m}}_g] = h_\alpha(\fvec, \gvec) \\
 \tilde{\fvec} = \fvec \circ \bm{\mathrm{m}}_f \quad \text{and} \quad \tilde{\gvec} = \gvec \circ \bm{\mathrm{m}}_g   \\
 [\theta,\bxx, \bxy] = ScanMatch(\tilde{\fvec}, \tilde{\gvec})
\end{gather}
where $\circ$ denotes the Hadamard product and $ScanMatch$ is defined in~\cref{alg:scan_match}. Given a dataset $\mathcal{D}=\{(\fvec, \gvec, \theta^*, \txx, \txy)_n\}_{n=1}^N$ the network parameters $\alpha$ are found by minimising:
\begin{equation}
\label{eq:loss}
    \mathcal{L}(\alpha) = \mathbb{E}_{\mathcal{D}} \left\{|\theta^* - \theta|_1 + |\txx - \bxx|_1 + |\txy - \bxy|_1\right\}
\end{equation}
Note that instead of minimising the Mean Square Error (MSE) as in \cite{barnes2019masking} we consider minimising the Mean Absolute Error (MAE) which is less sensitive to outliers. The network architecture for $h_\alpha$ is discussed further in \cref{sec:setup-network}.

\section{Experimental Setup}
\label{sec:Setup}
\subsection{Datasets}
\label{sec:offline}
We evaluate our approach using the \textit{Oxford Radar RobotCar Dataset}~\cite{barnes2020radarrobotcar} featuring a CTS350-X Navtech \acrshort{fmcw} radar with \SI{4}{\hertz} scan rate which defines our requirement for real-time.
In a similar approach to \cite{barnes2019masking,burnett2021radar,barnes2020under} we partition the data in \emph{time} rather than geography.
\cref{tab:splits} details the specific train, validation and test sets used.

\begin{table}[h]
\scriptsize
\centering
\vspace{-0.3em}
\begin{tabular}{llcc}
\toprule
    Split & Pattern & Examples & Percentage  \\
\midrule
    Train & \texttt{2019-01-1[1-8]*} & 197900 & 85$\%$ \\
    Validate & \texttt{2019-01-10-12-32-52*} & 8617 & 4$\%$ \\
    Test & \texttt{2019-01-10-1[24]*} & 25707 & 11$\%$ \\
\bottomrule
\end{tabular}
\caption{
All \textit{Oxford Radar RobotCar Dataset} loops which match the split pattern are used for each split.}
\label{tab:splits}
\vspace{-.4cm}
\end{table}

\subsection{Network Architecture And Training}
\label{sec:setup-network}
As our primary benchmark we compare against the \mbym~model proposed in \cite{barnes2019masking} which we train from scratch using the splits from \cref{sec:offline}. To ensure a fair comparison, the masking network architecture and masking strategy are kept consistent for both \mbym~and \mbympp~(see \cref{tab:network}).

\setlength\tabcolsep{1.5pt} 
\begin{table}[h]
\scriptsize
\centering
\begin{tabular}{ccccccccccccc}
\toprule
\multirow{2}{*}{In} & \multirow{2}{*}{Skip} & \multirow{2}{*}{Down} & \multicolumn{2}{l}{Conv} & \multirow{2}{*}{Norm} & \multirow{2}{*}{Act} & \multicolumn{2}{l}{Conv} & \multirow{2}{*}{Norm} & \multirow{2}{*}{Act} & \multirow{2}{*}{Up} & \multirow{2}{*}{Out} \\

                    &                       &                       & $c_i$       & $c_o$      &                       &                      & $c_i$       & $c_o$      &                       &                      &                     &                      \\
\midrule
\multicolumn{13}{l}{\textbf{Encoder}} \\
$\fvec, \gvec$        & $\noentry$            & $\noentry$            & 2           & 8          & $\bn$                 & $\relu$              & 8           & 8          & $\bn$                 & $\relu$              & -                   & $\feature_1$         \\
$\feature_1$        & $\noentry$            & $\maxpool$            & 8           & 16         & $\bn$                 & $\relu$              & 16          & 16         & $\bn$                 & $\relu$              & -                   & $\feature_2$         \\
$\feature_2$        & $\noentry$            & $\maxpool$            & 16          & 32         & $\bn$                 & $\relu$              & 32          & 32         & $\bn$                 & $\relu$              & -                   & $\feature_3$         \\
$\feature_3$        & $\noentry$            & $\maxpool$            & 32          & 64         & $\bn$                 & $\relu$              & 64          & 64         & $\bn$                 & $\relu$              & -                   & $\feature_4$         \\
$\feature_4$        & $\noentry$            & $\maxpool$            & 64          & 128        & $\bn$                 & $\relu$              & 128         & 128        & $\bn$                 & $\relu$              & -                   & $\feature_5$         \\
$\feature_5$        & $\noentry$            & $\maxpool$            & 128         & 256        & $\bn$                 & $\relu$              & 256         & 256        & $\bn$                 & $\relu$              & $\upsample$         & $\feature_6$         \\
\multicolumn{13}{l}{\textbf{Decoder}} \\
$\feature_6$        & $\feature_5$          & $\noentry$            & 384         & 128        & $\bn$                 & $\relu$              & 128         & 128        & $\bn$                 & $\relu$              & $\upsample$         & $\feature_7$         \\
$\feature_7$        & $\feature_4$          & $\noentry$            & 192         & 64         & $\bn$                 & $\relu$              & 64          & 64         & $\bn$                 & $\relu$              & $\upsample$         & $\feature_8$         \\
$\feature_8$        & $\feature_3$          & $\noentry$            & 96          & 32         & $\bn$                 & $\relu$              & 32          & 32         & $\bn$                 & $\relu$              & $\upsample$         & $\feature_9$         \\
$\feature_9$        & $\feature_2$          & $\noentry$            & 48          & 16         & $\bn$                 & $\relu$              & 16          & 16         & $\bn$                 & $\relu$              & $\upsample$         & $\feature_{10}$      \\
$\feature_{10}$     & $\feature_1$          & $\noentry$            & 24          & 8          & $\bn$                 & $\relu$              & 8           & 8          & $\bn$                 & $\relu$              & $\upsample$         & $\feature_{11}$      \\
$\feature_{11}$     & $\noentry$            & $\noentry$            & 8           & 2          &                       & $\sigmoid$           & $\noentry$  & $\noentry$ & $\noentry$            & $\noentry$           & $\noentry$          & $\bm{\mathrm{m}}_{fg}$ \\
\bottomrule
\end{tabular}
\caption{The network architecture $h_\alpha$ used to generate masks $\bm{\mathrm{m}}_f,\bm{\mathrm{m}}_g$ from radar scans $\fvec, \gvec$ in \cref{sec:approach-ours-learnt-embedding}. $\maxpool$ is max-pool, $\bn$ is batch-norm and $\upsample$ is for bi-linear upsampling.
}
\label{tab:network}
\end{table}
\setlength\tabcolsep{3.0pt} 

The scans $\fvec$ and $\gvec$ are concatenated to form a two channel tensor and passed to our network as a single input (adopting the best-performing \emph{dual} method from \cite{barnes2019masking}). A U-Net architecture~\cite{ronneberger2015u} is used to increase the feature dimension and decrease the spatial dimension through the repeated application of convolutions and max-pooling before this process is reversed through bi-linear up-sampling ($\upsample$) and convolutions \cite{odena2016deconvolution}. Information is allowed to flow from the encoder to the decoder using skip connections which are concatenated with the input feature map at each decoder level. Batch Norm ($\bn$) and ReLu activation ($\relu$) are applied after each convolution.  The masks $\bm{\mathrm{m}}_f,\bm{\mathrm{m}}_g$ output by our network are generated using a single convolution with a sigmoid activation.

As there is an intrinsic balance between run-time performance and input resolution with reference to~\cref{alg:scan_match} input parameters, we train both models at three resolutions $\delta_{xy} \in \{0.8, 0.4, 0.2 \}$ corresponding to input sizes $n_{xy} \in \{127, 255, 511\}$, similarly to \cite{barnes2019masking}, with a batch size of \num{128}, \num{64} and \num{32} respectively.
All networks are trained minimising the loss of \cref{eq:loss} for \num{80} epochs on the training set with no augmentation applied to the input data.
Translational drift (see \cref{sec:setup-odometry-success}) is calculated on the validation set at each epoch and the model with the smallest drift over all epochs is selected, before the accuracy is calculated over the \emph{test} set. We experimented with learning rates \num{1e-3} and \num{1e-4} using the Adam optimiser~\cite{kingma2014adam}, finding that all models perform best when training with a learning rate of \num{1e-4} with the exception of \mbympph~where \num{1e-3} was slightly better. For completeness we also include results which are available from the original implementation and splits, quoting directly from \cite{barnes2019masking}. We find that our implementation of \mbym~outperforms the original as presented in \cite{barnes2019masking} as shown in \cref{tab:kitti}. We attribute this to our introduction of batch-norms after every convolution, experimenting with slightly different resolutions ($127, 255, 511$ vs $125, 251, 501$) as well as a different training objective ($L1$ as opposed $L2$).
These observations may be useful when re-implementing our work and that of \cite{barnes2019masking}.

\subsection{Metrics}
\label{sec:setup-odometry-success}
To assess odometry accuracy we follow the KITTI odometry benchmark~\cite{geiger2012we}. For each \SI{100}{\metre} segment of up to
\SI{800}{\metre} long trajectories, we calculate the average residual translational and angular error for every test set sequence, normalising by the distance travelled. The performance across each segment and over all trajectories is then averaged to give us our primary measure of success.

As a core objective of this work, we also provide timing statistics using both a laptop without \acrshort{gpu} as well as an embedded device with limited graphics capability.
These test beds include a \textit{Lenovo ThinkPad} with Intel Core i7 \SI{2.9}{\giga\hertz} processor and \SI{8}{\giga\byte} \acrshort{ram} and a \textit{NVIDIA Jetson Nano} with a Quad-Core ARM Cortext-A57 \SI{1.42}{\giga\hertz} processor, \num{128} \acrshort{cuda} cores (\num{472} GFLOPS), and \SI{4}{\giga\byte} \acrshort{ram}.
During \textit{ThinkPad} and \textit{Jetson} tests, timing is measured by passing through the network tensors of batch size \num{1} which are populated by noise.
For \textit{Jetson}, we use event profiling provided by PyTorch/\acrshort{cuda}, while for \textit{ThinkPad}, we use the standard Python library.
All timing statistics stated are calculated by averaging between \num{2000} and \num{10000} forward passes.
We discard results from an initial ``burn-in'' of \num{50} to \num{100} steps in order to let computation stabilise.

\subsection{Baselines}
As our primary benchmark we compare our approach, \mbympp, against \mbym, as per \cite{barnes2019masking}.
Both models share the same masking network architecture and training setup (\cref{sec:setup-network}) and differ in how they solve for the pose (see~\cref{sec:Approach}).
We also include results for \mbym~and \mbympp~without masking, denoted as \raw~and \rawpp~respectively.
This allows us to further investigate the benefits that adopting a decoupled search brings to run-time performance.
Comparing \rawpp~to \mbympp~also allows us to compare our approach to a conventional decoupled procedure \emph{without} a learnt radar feature space, similar to~\cite{park2020pharao}.

\section{Results}
In \cref{results:run-time-performance}, \cref{results:accuracy} and \cref{results:training} we respectively investigate what impact a decoupled search has on run-time efficiency, real-time performance, and training.
In \cref{results:masking} we compare our approach with and without a masking network. 
Finally, in \cref{results:sparse} we investigate how our approach fairs in comparison to several sparse point-based baselines.

\label{sec:Results}

\subsection{Run-Time Performance}
\label{results:run-time-performance}

Comparing the run time efficiency of \mbympp~to \mbym~in \cref{tab:timings} the benefits of adopting a decoupled approach becomes clear; considering a like-for-like comparison at each resolution we are able to achieve speedups of \SIrange{372}{800}{\percent} on a \acrshort{cpu} and \SIrange{424}{470}{\percent} on the \textit{Jetson} (it is worth noting that the memory footprint of the \num{511} resolution \mbym~means it is unable to run on the \textit{Jetson} entirely).

\begin{table}[h]
\small
\centering

\vspace*{0.3cm}
\begin{subtable}[h]{0.45\textwidth}
\centering
\begin{tabular}{lcccccc}
\toprule
& \multicolumn{6}{c}{\textbf{Timing Results}} \\
& \multicolumn{3}{c}{\textbf{Think Pad (Hz)}} & \multicolumn{3}{c}{\textbf{Jetson (Hz)}}                                         \\
& 127    & 255             & 511            & 127 & 255 & 511 \\
\midrule
\textbf{Baseline}   &  & & & & \\
\texttt{mask}       & 96.2   & 33.4            & 7.6            & 24.7                    & 8.7                     & \multicolumn{1}{r}{2.4}       \\
\raw~  & 14.3   & 3.7             & 0.8            & 6.6                     & 2.1                     & -$^{1}$                             \\
\rawpp~& 83.2   & 58.2            & 21.0  & 28.3           & 22.3                    & \multicolumn{1}{r}{9.4}       \\
\mbym~      & 12.2   & 3.4             & 0.7            & 3.7                     & 1.4                     & -$^{1}$                             \\
&        &                 &                & \multicolumn{1}{l}{}    & \multicolumn{1}{l}{}    &                               \\
\textbf{Ours}       &        &                 &                & \textbf{}               & \textbf{}               & \multicolumn{1}{r}{\textbf{}} \\
\mbympp~    & 45.4   & 20.6   & 5.6   & 15.7           & 6.6                     & \multicolumn{1}{r}{1.9}  \\
\bottomrule
\end{tabular}
\caption{\label{tab:timings}}
\end{subtable}
\newline
\vspace*{0.2cm}
\newline
\begin{subtable}[h]{0.45\textwidth}
\centering
\begin{tabular}{lcccccc}
\toprule
& \multicolumn{6}{c}{\textbf{Kitti Odometry Error}}                           \\
& \multicolumn{2}{c}{127} & \multicolumn{2}{c}{255} & \multicolumn{2}{c}{511} \\
& Tra     & Rot          & Tra       & Rot         & Tra       & Rot         \\
\midrule
\textbf{Baseline} &       &             &           &             &     &    \\
\raw~     & 9.55      & 30.93     & 6.39      & 20.87      & 5.13      & 17.39      \\
\rawpp~   & 9.58      & 29.60      & 8.46      & 27.75      & 7.95      & 26.86      \\
\mbym~\cite{barnes2019masking}& 2.70     &  7.6      &  1.80     & 4.7      & 1.16      &  3.0     \\ 
\mbym~  & 2.15      & 6.46     & 1.36      & 3.98      &  -$^{2}$        &    -$^{2}$       \\
&           &             &           &             &           &             \\
\textbf{Ours}          &           &             &           &             &           &    \\
\mbympp~      & 2.77      & 8.74      & 2.01      & 6.3      & 2.00    & 6.3      \\  
\bottomrule
\end{tabular}
\caption{\label{tab:kitti}}
\end{subtable}
\caption{
Timing results (a) and Kitti Odometry Metrics (b). Timing results are in \SI{}{\hertz} while translational (Tra) and rotational errors (Rot) are in \SI{}{\percent} and \SI{}{\deg\per\kilo\metre} respectively. 
$^{1}$Failed to run entirely on the Jetson. 
$^{2}$Due to training time constraints and resource limitations values for \mbymh~ are not reported for our own re-implementation as the run-time performance of this model fell significantly below real-time as shown in \cref{tab:timings} (see \cite{barnes2019masking} for estimate).
}
\vspace{-.6cm}
\end{table}

Further insights into run-time efficiency are gained by considering the efficiency of the brute-force and decoupled scan matching procedure in isolation from the time taken to mask each radar scan. The former is determined by measuring the run-time performance of \mbym~and \mbympp~operating on \emph{raw} radar scan (without masking) and is given by \raw~and \rawpp~in \cref{tab:timings}. The latter is provided by measuring the time it takes for a forward pass through the masking network and is given by \texttt{mask}. Considering \raw~it becomes clear that the brute-force search for $\bxx, \bxy, \theta$ is a significant computational bottleneck; even without masking only the lowest resolution model is able to run in real-time ($>$\SI{4}{\hertz}, the radar scan rate) on the \textit{ThinkPad} and not at all on the \textit{Jetson}. In contrast the majority of \mbympp~models are currently throttled by the forward pass through the network, as can be seen by comparing \texttt{mask} to \rawpp~(where in the majority of cases the time taken for masking each radar scan is greater than that spent on the scan matching procedure).

\subsection{Real-Time Odometry Accuracy}
\label{results:accuracy}
As our approach runs faster we are able to use a model at a higher resolution whilst still maintaining real-time operation. 
Considering~\cref{tab:kitti}, we note that whilst increasing the resolution from $127$ to $255$ results in a significant reduction in end-point error we experience only a marginal reduction in error when increasing from a resolution $255$ to $511$ (e.g. \SI{2.01}{\percent} to \SI{2.00}{\percent}). As \mbymppm~runs significantly faster than \mbympph~we therefore consider \mbymppm~as our best performing model.

On the \textit{ThinkPad}, \mbymppm~outperforms the best performing (and only) real-time \mbym~model \mbyml~in terms of end-point error (\SI{2.01}{\percent}, \SI{6.3}{\deg\per\kilo\metre} vs. \SI{2.14}{\percent}, \SI{6.4}{\deg\per\kilo\metre}) whilst running \SI{168}{\percent} faster. 
For \textit{Jetson} tests \mbymppm~is still able to run in real-time at \SI{6.6}{\Hz}. This is in stark contrast to \mbym~which is unable to achieve \emph{real-time} performance at any of the tested resolutions.

\subsection{Training Comparisons}
\label{results:training}
\begin{figure}[h]
\centering
\includegraphics[width=\linewidth]{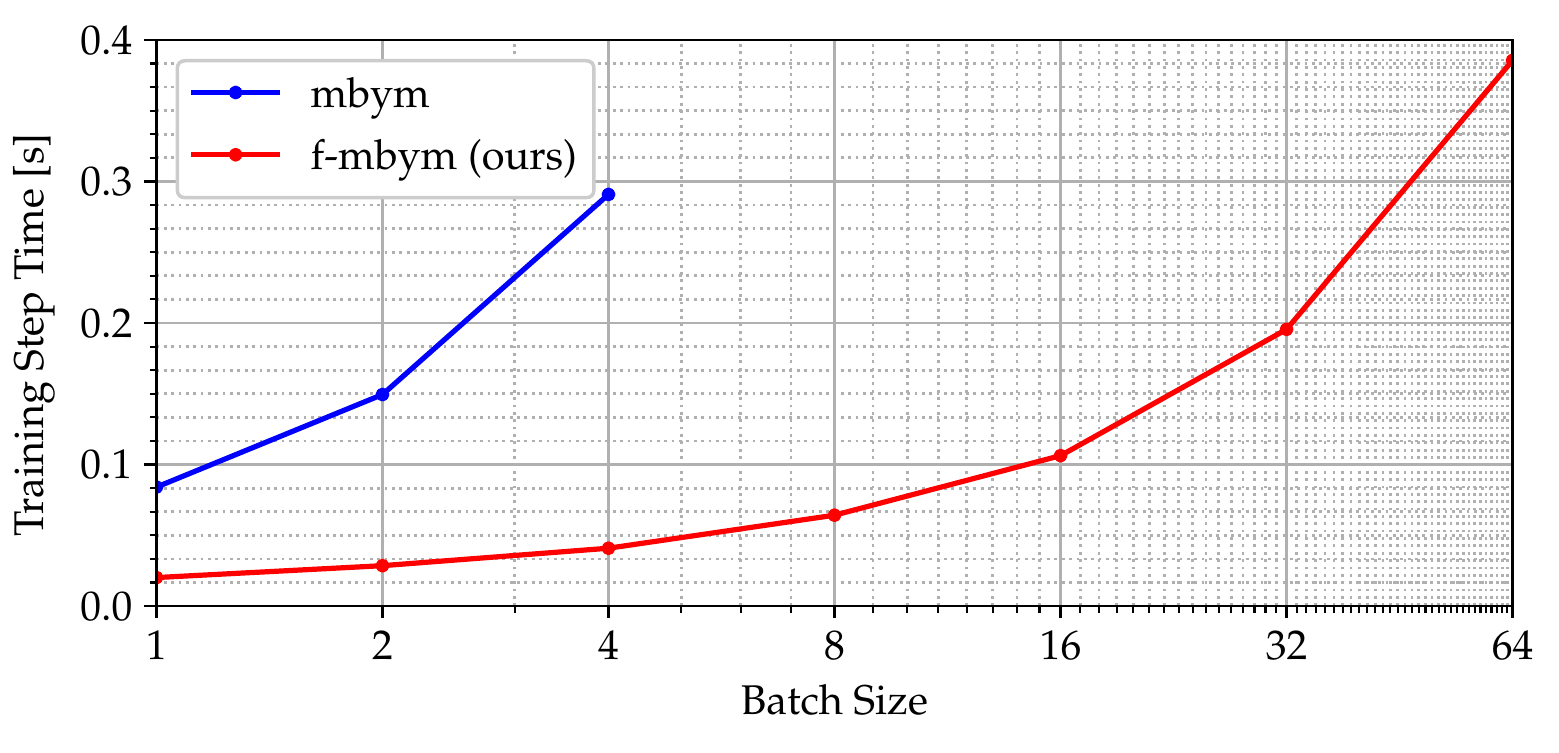}
\caption{Training step time comparison. Note that batch size is displayed using a log scale.}
\label{fig:training_time}
\vspace{-.6cm}
\end{figure}

By adopting a decoupled search for angle and translation we are able to train significantly faster and with much less memory. We average the time for each training step (excluding data loading) for \mbym~and \mbympp~running on $255$ resolution inputs across an epoch. This process is repeated, doubling the batch size each time, until a \textit{12GB Nvidia Titan X GPU} runs out of memory. The results are shown in \cref{fig:training_time}. Whilst \mbym~is only able to fit a batch size of \num{4} into memory, \mbympp~manages \num{64}. We also find that a training step for \mbympp~is $\sim{}4-7$ times faster than for \mbym~(a like-for-like comparison at each batch size).

\subsection{Masking}
\label{results:masking}
We now compare the performance of our approach with (\mbympp) and without (\rawpp) the masking network. Comparing the odometry accuracy (\cref{tab:kitti}) vs run-time performance (\cref{tab:timings}) of each method it is clear that increasing odometry accuracy is worth the added penalty to run-time performance. In the majority of cases \mbympp~is still able to run in real-time whilst increasing odometry accuracy by between \SIrange{345}{390}{\percent} across each resolution. We posit that conventional decoupled search approaches, as in \cite{park2020pharao} could experience similar boosts in performance by adopting a learnt feature representation as in our approach.

\subsection{Comparison To Sparse Point Based Methods}
\label{results:sparse}
Finally, we compare our approach to several existing point-based \gls{ro} systems on the \textit{Oxford Radar RobotCar Dataset}~\cite{barnes2020radarrobotcar}, including: Cen RO \cite{cen2018precise}, MC-RANSAC \cite{burnett2021we}, HERO \cite{burnett2021radar}, Under The Radar \cite{barnes2020under}, CFEAR \cite{adolfsson2021cfear}. For direct comparison we re-train our method using the splits from \cite{barnes2020under,burnett2021radar}. As shown in \cref{tab:kitti-other} we perform competitively with other approaches. We outperform Cen RO and MC-RANSAC by a significant margin. We also slightly outperform Under the Radar and HERO in rotational error. Only, CFEAR outperforms us in both translational and rotational error.

\begin{table}[h]
\small
\centering
\begin{tabular}{lcccc}
\toprule
& & \multicolumn{2}{c}{\textbf{Kitti Odometry Error}}                \\
\textbf{Method} & \textbf{Type} & Tra (\SI{}{\percent}) & Rot (\SI{}{\deg\per\kilo\metre}) \\
\midrule
\textbf{Sparse Point-Based} \\
Cen RO \cite{cen2018precise} & classical & 3.7168  & 9.50 \\
MC-RANSAC \cite{burnett2021we} & classical & 3.3190 & 10.93 \\
Under The Radar \cite{barnes2020under} & supervised  & 2.0583  & 6.70 \\
HERO \cite{burnett2021radar} & unsupervised & 1.9879 & 6.52 \\
CFEAR \cite{adolfsson2021cfear} & classical & 1.7600 & 5.00 \\
\\
\textbf{Dense} & & \\
\mbym \cite{barnes2019masking}& supervised & \emph{1.1600} & \emph{3.00} \\
\mbympp~(ours) & supervised & 2.0597 & 6.269 \\
\bottomrule
\end{tabular}
\caption{
Comparison to other recent \gls{ro} methods.
}
\label{tab:kitti-other}
\vspace{-.5cm}
\end{table}

\section{Conclusion}
\label{sec:Concl}
In contrast to the brute force search over all possible combinations of translation and angle proposed in \mbym~\cite{barnes2019masking}, we propose to decouple the search for angle and translation, exploiting the Fourier Transform's invariance to translation. Doing so allows our approach to be trained faster and with less memory as well as to run significantly faster at inference time.
By providing a greater run-time efficiency at higher resolutions our best performing real-time model achieves an end-point error of \SI{2.01}{\percent} in translation and \SI{6.3}{\deg\per\kilo\metre}, outperforming the best real-time Masking by Moving model in accuracy whilst running $168\%$ faster on a \acrshort{cpu} and in real-time (at \SI{6}{\hertz}) on a \textit{Jetson} GPU. Our approach is competitive with the current state of the art achieved by sparse, point-based methods, challenging the conventional wisdom that a sparse point-based method is necessary for real-time performance. 

As per \cref{results:run-time-performance} the run-time performance of our approach is currently limited by the time taken to mask each radar scan using a neural network. We also note that whilst our model achieves more accurate real-time performance in comparison to \cite{barnes2019masking} when considering a like-for-like comparison at each resolution a significant gap exists in odometry accuracy. Closing this gap further could allow a dense method to surpass the performance of a sparse method whilst running in real-time. Investigating whether this is achievable with the modifications to the proposed formulation alongside faster masking strategies constitute interesting areas for future research. Finally, the decoupled search developed in our approach, could also be used to efficiently search for larger rotations, and so utilised for metric localisation where the rotational offset can be arbitrary.

\section{Appendix}
\label{app:appendix}
As the affine transformation property of the \gls{ft} in ~\cref{eqn:affineft} is crucial to this work and the original description by Bracewell~\cite{bracewell1993affine} is not readily available, we derive it here again for completeness, starting with the definition of the 2D FT
\begin{align}
    \gt(\ww') &=\int_{\mathbb{R}^2} g(\xx') e^{- 2 \pi j {\ww'}^\top\xx'} d\xx' \label{eq:fft_def} \\
    &=\int_{\mathbb{R}^2} g(\Rx\xx+\tx) e^{- 2 \pi j {\ww'}^\top(\Rx \xx + \tx)} d\xx \label{eq:fft_subs} \\
    &=e^{- 2 \pi j {\ww'}^\top \tx} \int_{\mathbb{R}^2} g(\Rx\xx+\tx) e^{- 2 \pi j {\ww'}^\top\Rx \xx } d\xx \label{eq:fft_rearrange} \\
    &=e^{-2 \pi j (\Rx \ww)^\top {\tx} }\int_{\mathbb{R}^2} f(\xx) e^{- 2 \pi j \ww^\top \xx} d\xx \label{eq:u_def} \\
    &= e^{- 2 \pi j {\tx}^\top \Rx \ww} \ft(\ww) \label{eq:fft_def2}
\end{align}
\cref{eq:fft_subs} follows from a change of variables $\xx' = \Rx \xx + \tx$ noting $d\xx' = | \Rx |d\xx = d \xx$ and \cref{eq:fft_rearrange} by expanding the exponent and from the linearity of the Fourier transform. \cref{eq:u_def} follows by defining $\ww' = \Rx \ww$ and substituting $f(\xx) = g(\Rx \xx + \tx)$ as in \cref{sec:approach-formulation}. Finally, \cref{eq:fft_def2} follows from the definition of the 2D Fourier transform. Substituting $\ww' = \Rx \ww$ and rearranging terms finally gives \cref{eqn:affineft}:
\begin{align}
 \ft(\ww) &= \gt(\Rx \ww) e^{2 \pi j {\tx}^\top \Rx \ww} 
\end{align}


\section*{Acknowledgments}
This work was supported by EPSRC Programme Grant ``From Sensing to Collaboration'' (EP/V000748/1) as well as by the Assuring Autonomy International Programme, a partnership between Lloyd's Register Foundation and the University of York.
The authors would like to acknowledge the use of Hartree Centre resources and the University of Oxford Advanced Research Computing (ARC) facility in carrying out this work {\tt\small http://dx.doi.org/10.5281/zenodo.22558}.
We gratefully acknowledge our partners at Navtech Radar and the support of Scan UK in this research.

\newpage
\bibliography{bibliography}
\bibliographystyle{ieeetr}

\end{document}